# Instantaneously Trained Neural Networks

Abhilash Ponnath

**Abstract:** This paper presents a review of instantaneously trained neural networks (ITNNs). These networks trade learning time for size and, in the basic model, a new hidden node is created for each training sample. Various versions of the corner-classification family of ITNNs, which have found applications in artificial intelligence (AI), are described. Implementation issues are also considered.

## 1 Introduction

The human brain, the most complex known living structure in the universe, has the nerve cell or neuron as its fundamental unit. The number of neurons and connections between the neurons is enormous; this ensemble enables the brain to surpass the computational capacity of supercomputers in existence today. Artificial neural networks (ANNs) are models of the brain, which implement the mapping, $f: X \rightarrow Y$ such that the task is completed in a "certain" sense. These networks are able to learn training samples as well as generalize them, which makes them an interesting research area. Unfortunately, it is doubtful that current ANN models will ever have learning capacity that will match the performance of living systems. Furthermore, the number of iterations required for generalization is often excessive and this motivates us to look into network designs where learning is fast.

For comparison with biological systems, it should be noted that memory is not stored in a single area of the brain but distributed. Also, different parts of the brain are required for processing different kinds of memories. The hippocampus, parahippocampal region and areas of the cerebral cortex (including prefrontal cortex) support declarative or cognitive memory and the amygdala, striatum, and cerebellum support the non-declarative or behavioral memory. Declarative knowledge can be classified into working, episodic, semantic memories and requires processing in the medieval temporal region and parts of the thalamus (Fig. 1). Non-declarative knowledge requires the processing of the basil ganglia. Semantic memory represents the individual's knowledge in general, whereas episodic memory structures individual's experiences.

Creating ANNs that can learn and generalize information seems to be the first step in the design of connectionist intelligent machines, and this is done by several ANN models. However, the learning and generalization time for the most popular models can be very large. The ability to store information quickly not only provides a capacity that is common in biological systems, it has obvious applications in computational systems.

Scholarly opinion is divided on whether AI machines can have real intelligence. It is not clear why complexity alone in connectionist [17-19] or rule-based [6,20-22,26,30] machines would suddenly create self-referring intelligence [14]. If classical computing models (and these include ANNs) are deficient, then such intelligence may be a consequence of quantum information processing in the brain, but current models of quantum information processing suffer from their own shortcomings [e.g.9,10,12,13,15], and we will not consider them in this paper (we list them here to give a flavor of the issues involved both in respect of speed of computation as well as storage of information).

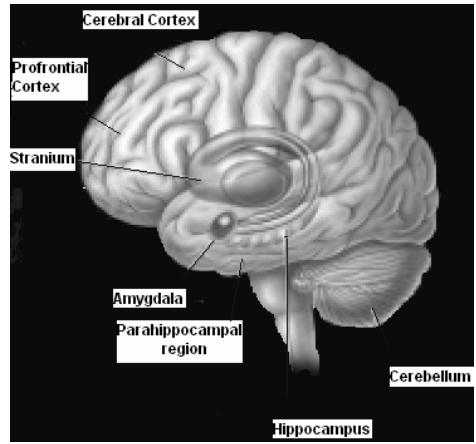

Fig 1: Different regions of the brain

This paper is written to provide an introduction to the field of instantaneously trained neural networks (ITNN), which matches the biological capacity of fast learning. The paper is organized as follows: Section 2 provides a brief historical background to research in neural networks. ITNNs are described in Section 3, applications are presented in Section 4 and hardware implementations of a specific type of ITNN are presented in Section 5.

**2 Neural network models**

The first model of a neuron was proposed in 1943 by Warren McCulloch and Walter Pitts, who showed that a network built with sufficient number of the neurons with proper weights, was capable of any feasible computation. The network suffered from the limitation that it lacked the capacity to learn. D.Hebb gave the first neural network with learning capability [3], based on the correlation principle that if neuron x was repeatedly stimulated by neuron y at times neuron x is active, then neuron x will become more sensitive to stimuli from neuron y. This lead to the notion of adjustable synaptic weights which is incorporated into most neural networks we know today. Frank Rosenblatt proposed a class of neural networks called *perceptrons*.

It was believed that neural networks could perform any function until 1969 when Minsky and Papert pointed out certain computations limitations of the perceptrons [21]. Specifically, they showed that they could not perform tasks like Exclusive-XOR. The field lay dormant until mid 80s when the *backpropagation* (BP) algorithm was introduced. Considered a breakthrough in area of neural networks, this algorithm suffers from the drawback that it may not converge [1,16,19]. The ADALINE (ADAptive LInear NEuron) was invented by Widrow and his group. An adaptive learning algorithm was used to adjust the weights by minimizing the mean square error. MADALINE was consisted of many ADALINE networks in parallel. The output was either 1 or 0 depending on the output from individual ADALINE units.

Networks without a feedback path are called *feedforward* networks .Two different configurations exist in this category they are single-layer feedforward networks and multi-layer feedforward networks.

## 2.1 Radial basis function networks

Radial basis function (RBF) networks have their roots in older pattern recognition techniques such as functional approximation and mixture models [19]. The RBF is a two layer neural network in which each hidden unit implements a radial activated function. The output units implement a weighted sum of the hidden unit outputs. RBFs possess nonlinear approximation properties and hence can model complex mappings. The mapping for the input is nonlinear and linear to the output. The commonly used RBFs are Gaussian, piecewise-linear approximation, cubic approximation, multiquadratic function, and inverse-multiquadratic function. RBF's convergence of weights is faster and less sensitive to interference compared to a BP network.

Neural networks have two important capabilities: (a) learning (b) generalization. The task at hand for a neural network is to learn a model in which it is embedded; observations are made and pooled to form the training samples to train the network. The observations pooled are either labeled or unlabeled depending weather the sample is an input-output pair or not. Learning can be classified as supervised or unsupervised depending weather the sample is labeled or unlabeled.

## 2.2 WIZARD

WIZARD consists of RAM chips which has look up tables which implement the neuron functions. The training is done by updating the contents of the RAM chips.  The input and output must be digitized into binary vectors hence a back draw of this network and cannot be used for time series prediction. The schematic diagram of a RAM node is shown below. The node performs logical functions and returns a value to the given input vector.

A discriminator is formed by arranging a group of k units .The output of a discriminator is the sum of the individual RAM nodes. The network consists of discriminators in parallel with each discriminator trained to recognize a different class of pattern. Initially all the $k2^N$ locations of the discriminator are set to zero and training is done by applying a input pattern at the input and the RAM nodes are set to value 1.

If the same input pattern is presented to the discriminator, the RAM node is set to read mode and previously written locations will all be accessed and RAM nodes will be responded with 1. If a noisy version is presented then the output will be a function of number of previously written locations that are accessed by the noisy input. This is proportional to the similarity of the input pattern without noise to input pattern with noise.

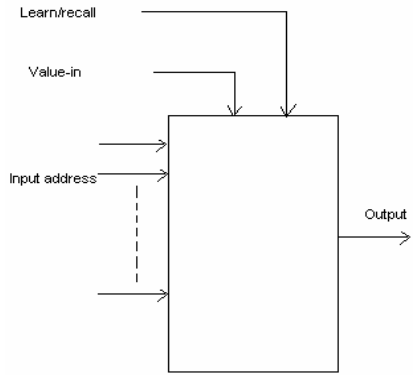

Fig 2: A RAM node

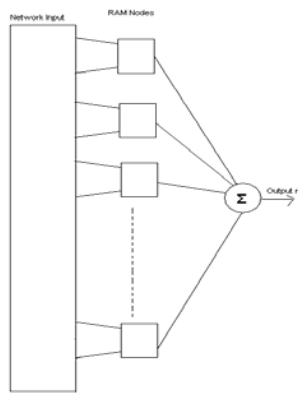

Fig 3: A discriminator

### 2.3 Probabilistic neural network (PNN)

A probabilistic neural network (PNN) may be viewed as a normalized RBF network in which there is a hidden unit for each training value [27]. These hidden units are called "kernels" and they are typically probability density functions such as the Gaussian. The hidden-to-output weights are usually 1 or 0; for each hidden unit, a weight of 1 is used for the connection to the output which is true for it, and all other connections are given weights of 0. The other possibility is to assign weights for the prior probabilities of each class. The weights that need to be learned are the widths of the RBF units.

PNN training is not iterative in the same sense as backpropagation, but it requires estimation of kernel bandwidth, and it may require the processing of all training information. In reality, therefore, it is not instantaneous.

The principle of operation of PNN is based on statistical technique, combining both the bayes strategy and a nonparametric estimation technique. The estimator has a Gaussian probability distribution function:

$$f_k(x) = \left[(2\Pi)^{R/2} \sigma^R S\right]^{-1} \sum_{i=1,S} \exp\left(\frac{-(X-X_{ki})^T(X-X_{ki})}{2\sigma^2}\right)$$

A PNN for a three-class problem is now described. The inputs are R-dimensional continuous valued vectors normalized to unit length. The input units receive the inputs and feed the to the pattern units, each of which form a dot product with a weight vector and perform a nonlinear operation on the dot product and feeds to the summation unit its activation level. The nonlinear operation performed is an exponential function of the form $\exp[(z_i-1)/\sigma^2]$. $\sigma$ is called smoothing parameter. For good performance $\sigma$ takes values between 0.2 and 0.3 [3].

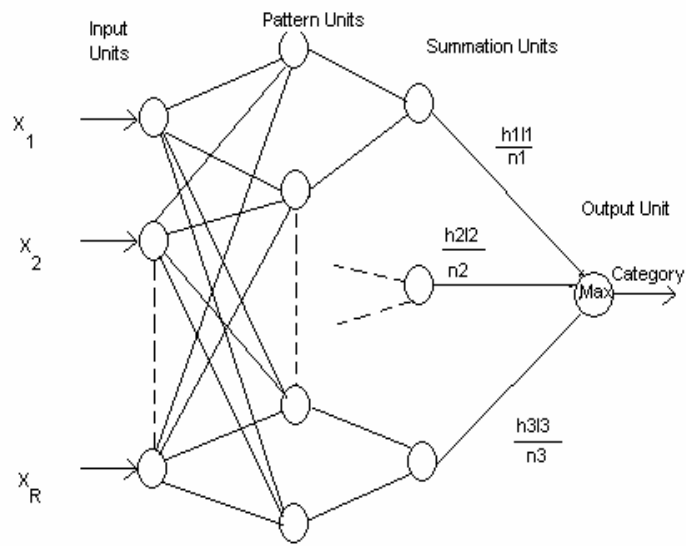

Fig 4: A PNN with 3 output classes

The weight vector in each of the pattern is set equal to one of the vectors in the training set and then connecting the pattern unit output to the appropriate summation unit with a connection weight of one. Using Bayes decision theory a test pattern can be assigned to category k if

$$h_k l_k f_k(x) > h_q l_q f_q(x) \quad \forall q \neq k$$

$h_k$ is a priori probability of occurrence of patterns from category k
$l_k$ is loss associated with classifying a test pattern into category other than k when in reality belongs to k. The output unit receives the output from each summation multiplied by $\frac{h_k l_k}{n_k}$ and then determines the category into which the vector must be classified. This network has the drawback that it cannot be used for applications involving function approximation. We need to use the generalized regression neural networks (GRNN) if the application involves function approximation.

We note that WIZARD and PNN do not possess instantaneous learning capabilities although that is claimed for them by some people,

## 3 Instantaneous neural networks

The principal ITNN is the corner classification neural network (CCNN) that includes its variant FC neural network (FCNN). These networks are an attempt to model biological memory. The FC networks are not purely instantaneous, and they need some learning. However, this learning could be done very quickly under certain conditions. Their generalization capacity seems to be almost as good as that of backpropagation networks [23, 28, 29].

Memory may be divided into three types, sensory, short-term and long-term memory. The duration for which information can be retained is shortest for sensory memory and longest for greatest for long-term memory, short-term memory stands in between the sensory and long-term memory.

*Short-time memory* also called working memory involves the ruminate thoughts we have just encountered, the information fades approximately after twenty seconds if it is not renewed through rehearsal. Short-term memory needs to be protected from overloading by sensory stimulation, two cognitive processes that help in preventing overloading are s*ensory gating* and *selective attention*. *Sensory gating* is the process by which certain channels are turned on while others are turned off. *Selective attention* is the process of culling information received by one channel i.e. abate information 'x' received by a channel in favor to information 'y' entering the same channel. The amount of information that short-term memory can hold is limited but it can be extended by "grouping" information.

### 3.1 Corner classification neural networks (CCNN)

The corner classification (CC) network is based on the idea of phonological loop and the visio-spatial sketchpad [2,11]. It was proposed by Kak in 1992 in three variations [7,8]. These and its more advanced variants are also known as the class of Kak neural networks.

The concept of radius of generalization was introduced in CC3 and thus this neural network overcame the generalization problem that plagued the earlier CC2 network. The Hamming distance was used for classification between binary vectors, i.e. any test vector whose Hamming distance from a training vector is smaller than the radius of generalization of the network is classified in the same output class as that training vector. A unique neuron is associated with each training sample and each node in the network acts as a filter for the training sample. The filter is realized by making it act as a hyper plane to separate the corner of the n-dimensional cube represented by the training vector and hence the name corner-classification (CC) technique.

There are four versions of the CC technique, represented by CC1 through CC4. The CC4 is shown to be better than the other networks in the CC category [5]. The number of input and output neurons is equal to the length of input and output patterns or vectors. The number of hidden neurons is equal to number of training samples the network requires. The last node of the input layer is set to one to act as a bias to the hidden layer. The binary step function is used as activation function for both the hidden and output neurons. The output of the function is 1 if summation is positive and zero otherwise.

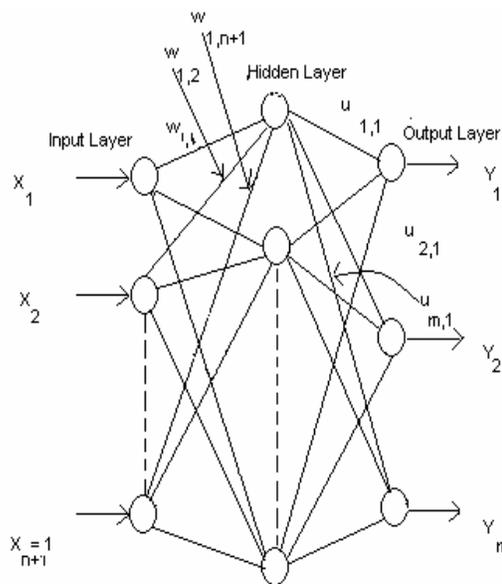

Fig 5: General CC4 architecture

The hidden neuron receives a -1 if the input vector is $-1$ and 1 if the input is 0 and +1. The weight of the link from base node to a hidden neuron is $r-s+1$, r is the radius of generalization and s is the number of ones in the input sequence. The weights in the output layer are equal to 1 if the output value is 1 and $-1$ if the output value is 0. This amounts to learning both the input class and its complement and thus instantaneous. The radius of generalization, r can be seen by considering the all-zero input vectors for which $w_{n+1} = r + 1$. The choice of r will depend on the nature of generalization sought. This network also suffers from the draw back that the input output data must be digitized.

**3.2 FC neural network**

Human decision making has the element of uncertainty in it, for example when we interpret temperature we can only express the degree of hotness or coldness, this is not like a truth statement which might be true or false. One might say it is hot, another might say its or very hot etc. The principle of fuzzy classification networks depend on the concept of "nearest neighbor", it consists of three layers-an input layer, a hidden layer and an output layer. The FC acronym has been seen either to stand for fuzzy classification or fast classification [29].

The input data is normalized and presented as input vector **x**. The hidden neuron is represented by the weight vector **w$_i$** and its elements are represented by $w_{i,j}$, i=(1,2,…S) and j=(1,2,…,R). The output is the dot product of the vectors **μ** and **u**. This network can be trained with just two passes of the samples, the first pass assigns the synaptic weights and the second pass determines the radius of generalization for each training sample. By fuzzification of the location of the each training sampler and by assigning fuzzy membership functions of output classes to new input vectors.

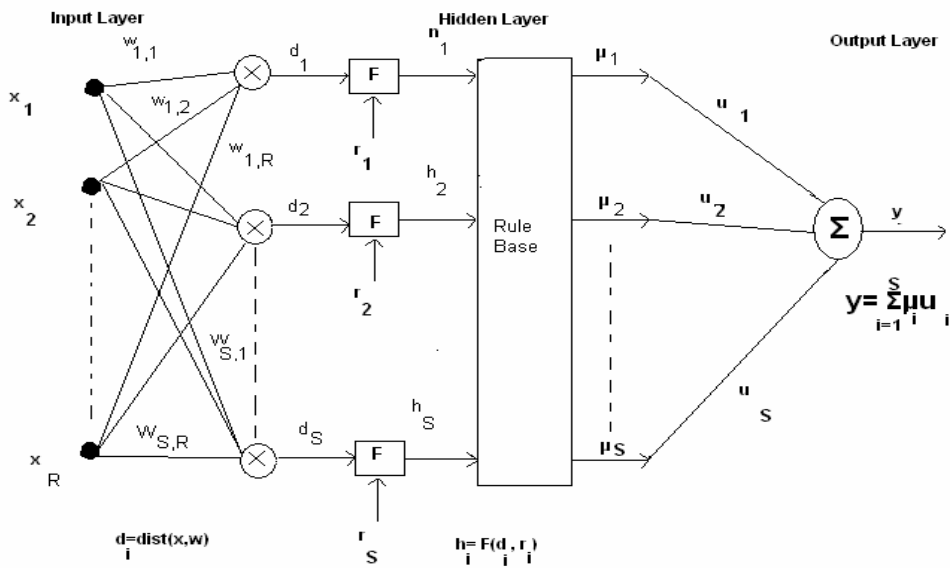

Fig 6: FC network architecture

The network behaves as a 1NN classifier and a kNN classifier according to weather the input vector falls within the radius of generalization of a training vector or not hence the radius of generalization acts as a switch between the 1NN classifier and the kNN classifier. The FC network meets the specifications set by traditional function approximation that every data point is covered in the given training sample and also Cover's theorem on separability of patterns. In practical case k values is determined by the sample size and be a fraction of the sample size. If k=S then the FC network operating as a kNN classifier can be viewed as a RBF network provided the membership function is chosen to be a Gaussian distributed, moreover if the weighting function is chosen to be the membership function the FC network can be considered as a kernel regression.

## 4 Applications of ITNNs

We speak of three broad applications that may be listed as (1) function approximation (2) time series prediction (3) pattern classification and data fusion for metasearch engines. Each of these areas naturally covers many specific applications. For example, time series prediction may be applied to financial or signal data, and pattern recognition includes application to engineering, military, or medical systems.

### 4.1 Function approximation

The problem in function approximation is to implement a function, F(x), which approximates an unknown function, f(x), with a priori as the input-output pairs in a Euclidian sense for all inputs.

The input-output pairs form the training samples. The error can be made as small as possible by increasing the size of the sample space.

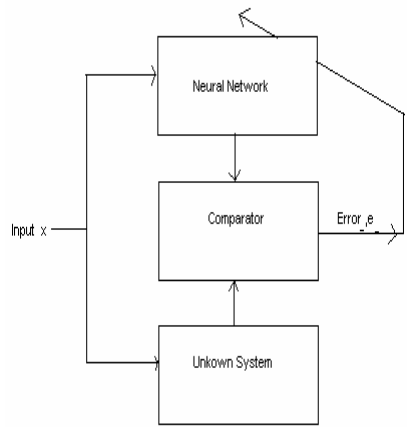

Fig 7: System identification

*System Identification and Inverse system* modeling are two areas where ITNNs can be useful. The schematic block diagrams for both the areas are shown in Figures 7 and 8 along with the neural network block.

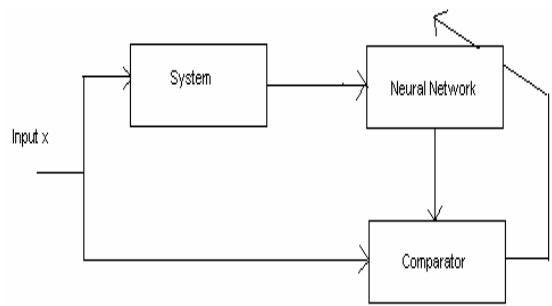

Fig 8: Inverse system modeling

Specific applications in representation and control of systems need to be investigated further.

### 4.2 Time series prediction

Time series prediction is widely used in financial data for prediction of stocks, currency and interest rates and engineering such as electric load demand. The network is trained by historical data set with time index and the network predicts the future values based on past values. Two time series examples that have been investigated are Henon map and Mackey-Glass time series [28,29], which are both chaotic time series and, therefore, their statistical properties remain unchanged.

We don't know how ITNNs perform for real world data, which is characterized by different modes, as compared to other neural network techniques.

### 4.3 Pattern classification

Our brain is highly developed and can recognized different visual, audio and olfactive patterns. Neural networks can be trained in order to recognize the patterns. The term "pattern classification" refers to the process by which the network learns from a given set of training values the mapping between the given input-output pairs and attempts to assign new input patterns to one of the output classes predefined by the training classes. Studies show that ITNNs do almost as well as backpropagation networks in basic pattern classification problems [23,28].

Metasearch engines combine results to queries submitted to different search engines after discarding redundant results. ITNNs can be used to fuse data in this application. In a method proposed by Shu and Kak [25] the network was built using the keywords from web pages. Individual key words were set either to 1 or 0 depending weather they were in top or bottom of a list of search results.

## 5 Implementation of ITNNs

Implementation of an algorithm speaks of its success in the commercial arena. The CC4 algorithm was implemented using reconfigurable computing and to design an optical neural network .The FC network was implemented on FPGAs.

Zhu and Sutton [5] implemented Kak's FC network on FPGAs. The Celoxica RC2000 board with Xilinx XC2V6000 Virtex-II chip was used and JHDL hardware description language was used for design and simulation for hardware implementation. The hidden neuron circuit was based on Euclidian distance in Kak's FC network but such a implementation in a FPGA would require a computational complexity of O(mn) "square operations" where m is the number of neurons and n is the number of elements in the weight vector. Hence, the two different distance metrics were proposed, the city block distance and the box distance. It should be noted that a general distance metric is defined as $d_i = \|x_i - w_i\|^p$ and the value of the parameter p gives the distance different names, as shown below for case p=1, 2, ∞.

$$d_i = \begin{cases} \|x_i - w_i\|^1 = \sum_{j=1}^{n} |x_{i,j} - w_{i,j}|^1 & p = 1 \quad \text{the city block distance} \\ \|x_i - w_i\|^2 = \sum_{j=1}^{n} |x_{i,j} - w_{i,j}|^2 & p = 2 \quad \text{the Euclidian distance} \\ \|x_i - w_i\|^\infty = Max(|x_{i,j} - w_{i,j}|) & p = \infty \quad \text{the box distance} \end{cases}$$

The use of the two different distance metrics is justified by the experiments done by Estlick which gave acceptable results [1]. The activation circuit was implemented as an n-bit constant comparator as the radius of generalization is constant after training. Bitonic selection network is used to implement the kNN circuit. Implementation on FPGA requires that the algorithm be simple, modular and highly parallel and the FC networks possess these characteristics. The ease of implementation of the FC networks on a FPGA justifies the claim that ITNNS can be modeled into hardware and can be commercialized.

Zhu and Milne [31] showed that Kak's CC4 is hardware implementable in reconfigurable computing using fine grained parallelism. Shortt, Keating, Moulinier, Pannell [24] made an optical implementation the Kak neural network using a bipolar matrix vector multiplier, but suitable modifications to the structure and training algorithm were required to build an optical neural network implementing *N*-parity.

## 6 Conclusions

This paper provides a review of neural networks with an emphasis on ITNNs. The corner classification network learns by creating a new hidden node for each training sample. Its advantage in speed (for training) is a consequence of the price that is being paid in size of the network. While these hidden neurons can be pruned, the size will remain much larger than that of a backpropagation network. In applications, where the size does not matter, these networks are an attractive alternative to backpropagation. Although, they have been evaluated for various kinds of chaotic time series, the performance of these networks on various benchmark data remains to be checked.